\documentclass[10pt,twocolumn,letterpaper]{article}

\usepackage{cvpr}              % To produce the CAMERA-READY version
\usepackage{graphicx}
\usepackage{amsmath}
\usepackage{amssymb}
\usepackage{booktabs}

\usepackage[pagebackref=false,breaklinks=false,letterpaper=true,colorlinks,urlcolor=blue,citecolor=blue,linkcolor=blue,bookmarks=false]{hyperref}

\usepackage[capitalize]{cleveref}
\crefname{section}{Sec.}{Secs.}
\Crefname{section}{Section}{Sections}
\Crefname{table}{Table}{Tables}
\crefname{table}{Tab.}{Tabs.}

\def\cvprPaperID{3709} % *** Enter the CVPR Paper ID here
\def\confName{CVPR}
\def\confYear{2023}

\begin{document}

\title{\vspace{-16mm}SinDiffusion: Learning a Diffusion Model from a Single Natural Image \vspace{-6mm}}

\author{Weilun Wang{\small $~^{1}$}, ~Jianmin Bao{\small $~^{2}$}\thanks{Corresponding author: Jianmin Bao.}, ~Wengang Zhou{\small $~^{1,4}$}, \\
Dongdong Chen{\small $~^{3}$}, ~Dong Chen{\small $~^{2}$}, Lu Yuan{\small $~^{3}$}, ~Houqiang Li{\small $~^{1,4}$}\\
\normalsize
$^{1}$ CAS Key Laboratory of GIPAS, EEIS Department, University of Science and Technology of China (USTC) \\
\normalsize
$^{2}$ Microsoft Research Asia \enspace \enspace \enspace $^{3}$ Microsoft Cloud+AI\\
\normalsize
$^{4}$ Institute of Artificial Intelligence, Hefei Comprehensive National Science Center\\
\normalsize
{\tt\small wwlustc@mail.ustc.edu.cn, \{jianbao,dochen,doch,luyuan\}@microsoft.com, \{zhwg,lihq\}@ustc.edu.cn}
}

\twocolumn[{%
\renewcommand\twocolumn[1][]{#1}%
\maketitle
\begin{center}
\centering
\vspace{-12mm}
\includegraphics[width=\textwidth]{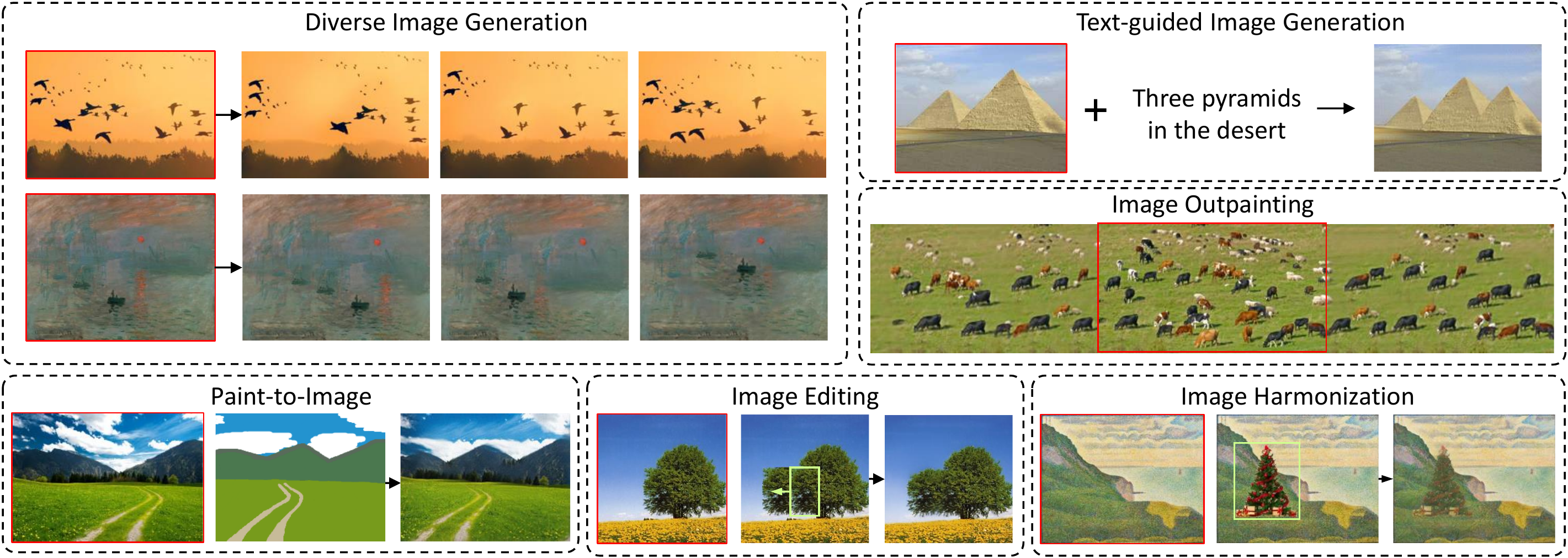}
\vspace{-7mm}
\captionof{figure}{
\textbf{Image generation and manipulation results on SinDiffusion.}
Images with red border denote the training images.
SinDiffusion generates realistic and diverse images by learning from a single natural image, \emph{e.g.}, landscapes and arts.
Meanwhile, SinDiffusion is applicable to various downstream applications and produces high-fidelity results which correspond with the condition.
}
\label{fig:teaser}
\end{center}%
}]

%%%%%%%%% ABSTRACT
\begin{abstract}
\vspace{-3.4mm}
We present SinDiffusion, leveraging denoising diffusion models to capture internal distribution of patches from a single natural image. 
SinDiffusion significantly improves the quality and diversity of generated samples compared with existing GAN-based approaches. It is based on two core designs. First, SinDiffusion is trained with a single model at a single scale instead of multiple models with progressive growing of scales which serves as the default setting in prior work. This avoids the accumulation of errors, which cause characteristic artifacts in generated results. Second, we identify that a patch-level receptive field of the diffusion network is crucial and effective for capturing the image’s patch statistics, therefore we redesign the network structure of the diffusion model. Coupling these two designs enables us to generate photorealistic and diverse images from a single image. Furthermore, SinDiffusion can be applied to various applications, \emph{i.e.}, text-guided image generation, and image outpainting, due to the inherent capability of diffusion models.
Extensive experiments on a wide range of images demonstrate the superiority of our proposed method for modeling the patch distribution. 
The code and models are publicly available at \url{https://github.com/WeilunWang/SinDiffusion}.
\footnotetext{*Corresponding author: Jianmin Bao.}
\end{abstract}

%%%%%%%%% BODY TEXT
\vspace{-1mm}
\section{Introduction}
\label{sec:intro}
Generating images from a single natural image has extracted more and more attention due to its various applications.
This task aims to learn an unconditional generative model from a single natural image to generate diverse samples with similar visual content by capturing the internal statistics of patches.
Once trained, the generative model can not only produce high-quality diverse images of arbitrary resolutions but also easily adapt to multiple applications, \emph{i.e.}, image editing, image harmonization, and image-to-image translation.

\begin{figure}[t]
  \centering
   \includegraphics[width=\linewidth]{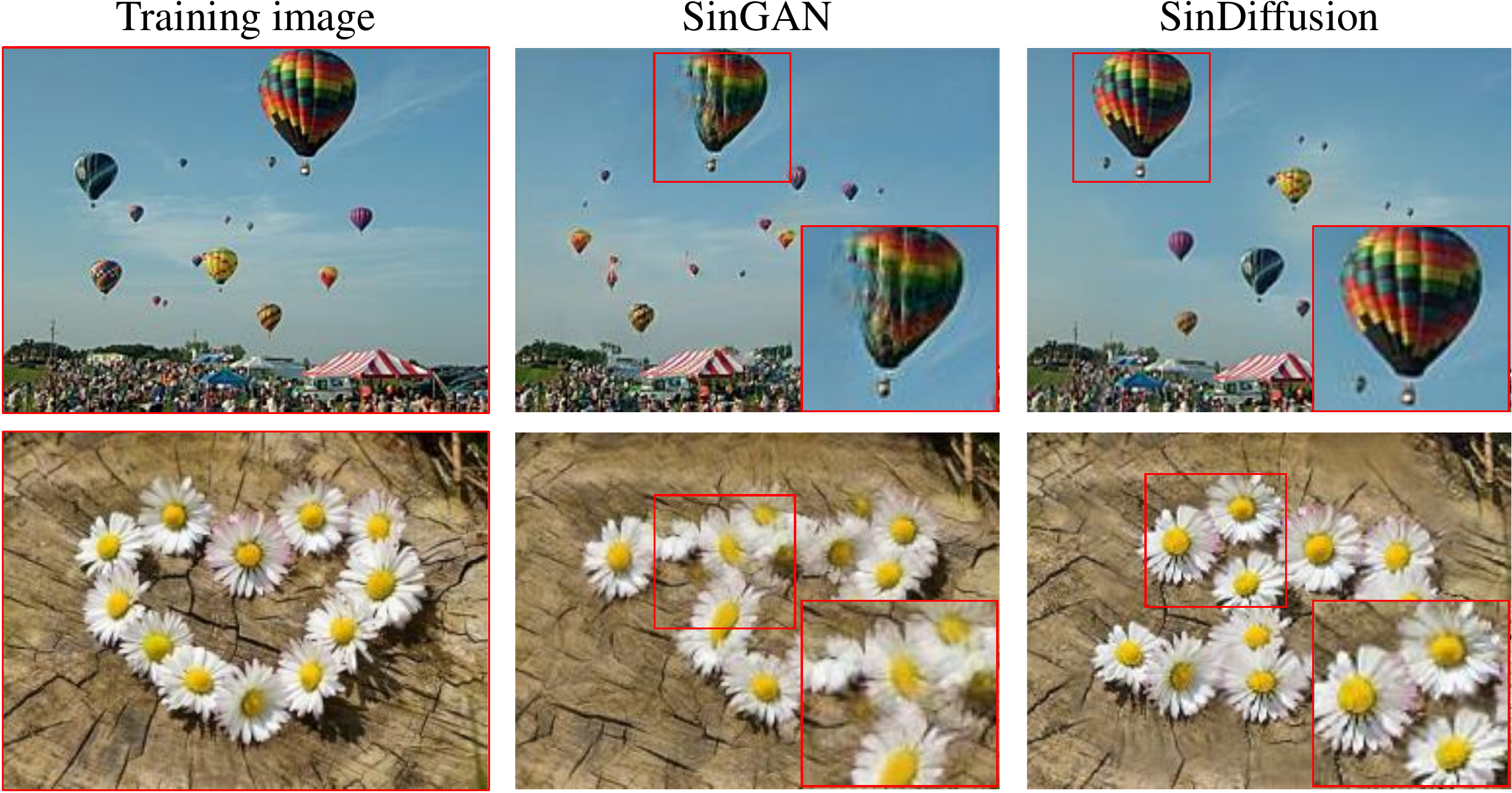}
   \vspace{-6mm}
   \caption{
   Images generated by SinGAN and SinDiffusion.
   It is observed that the images generated by SinGAN contain blurred textures and characteristic artifacts, especially at the boundaries of objects.
   Different from that, SinDiffusion generates realistic images which semantically resemble the training image.
   }
   \vspace{-5mm}
   \label{fig:artifact}
\end{figure}

The groundbreaking method of this task is SinGAN~\cite{shaham2019singan}, which builds multiple scales of the natural image and trains a series of GANs to learn the internal statistics of patches in a single image. The core idea of SinGAN is to train multiple models at progressive growing scales. This becomes the default setting of this direction.
However, we may observe these methods generate unsatisfactory images. This is because these methods accumulate the small detail errors produced at small scales, which then lead to obvious characteristic artifacts in resulted images (see Figure~\ref{fig:artifact}).

In this paper, we propose a novel framework called Single-image Diffusion Model~(SinDiffusion) for learning from a single natural image. SinDiffusion is based on the recently developed Denoising Diffusion Probabilistic Model (DDPM). We find that multiple models at progressive growing scales are not essential for learning from a single image, 
%Unlike previous methods that employ multiple models at progressive growing scales, SinDiffusion tackles this problem with a single denoising diffusion probabilistic network at a single scale.
A single diffusion-based model trained on a single scale is well-suited for this task. Although the diffusion model is a multiple-step generation process, it doesn't suffer from the issue of accumulated errors. The reason is that the diffusion model has systemically mathematical formulations and the error in intermediate steps can be regarded as noise which could be refined during the diffusion process. 

The other core design of SinDiffusion is to restrict the receptive field of the diffusion model.
We revisit the commonly-used network structure in the previous diffusion model~\cite{dhariwal2021diffusion} and find it enjoys a strong capability with a deep structure. The network structure has a large receptive field that can cover the full image, which leads to the model tending to memorize the training image and thus generating exactly the same image as the training image.
To encourage the model to learn the patch statistics rather than memorize the whole image, we carefully design the network structure and introduce a patch-wise denoising network.
Compared with the previous diffusion structure, SinDiffusion reduces the downsampling times and the number of resblocks in the original denoising network structure. Equipped with this design, SinDiffusion produces high-quality and diverse images by learning from a single natural image (see Figure~\ref{fig:artifact}).

Our proposed SinDiffusion enjoys the advantages of flexibility for various applications~(see Figure~\ref{fig:teaser}). It can be used for various applications without any re-training of the model.
%In addition, as a kind of score-based model, diffusion models are inherently capable of solving a variety of image manipulation tasks with the unconditional model.
In SinGAN, the downstream applications are mainly implemented by feeding the condition into the different scales of pre-trained GANs.
Therefore, the applications of SinGAN are limited to those given ``spatially-aligned'' conditions.
Different from that, SinDiffusion is available for a wider range of applications by designing the sampling procedure.
SinDiffusion learns to predict the gradient of the data distribution through unconditional training.
Supposing a score function~(\emph{i.e.}, $L-p$ distance or a pre-trained network like CLIP~\cite{radford2021learning}) which describes the relevance between the generated image and condition, we utilize the gradient of the relevance score to guide the sampling procedure of SinDiffusion.
In this way, SinDiffusion is able to generate images that both fit the data distribution and are corresponding to the given conditions.

To demonstrate the superiority of our proposed framework, we conduct experiments on a variety of natural images, including landscapes and famous art.
Both quantitative and qualitative results validate that SinDiffusion can generate both high-fidelity and diverse results.
Downstream applications further demonstrate the usefulness and flexibility of our SinDiffusion.

Overall, the contributions are summarized as follows,
\begin{itemize}
    \item We propose a novel diffusion-based framework named \textbf{SinDiffusion} for capturing the internal statistics of patches from a single natural image.
    \vspace{-2mm} %\item We find that a single-scale patch-wise denoising network is well-suited for this task, which can generate more photorealistic and diverse images compared with previous GAN-based methods.
    \item We introduce two key ingredients in SinDiffusion: single-scale training and a new network with patch-level receptive fields, techniques essential for generating high-quality and diverse images.
    \vspace{-2mm} \item We explore more downstream applications by leveraging the advantages of SinDiffusion, including text-guided image generation, image outpainting, and \emph{etc.}
    \vspace{-6.1mm} \item Extensive experiments on various natural images, \emph{i.e.}, landscapes and famous arts, demonstrate the effectiveness and wide applicability of our framework.
\end{itemize}

%-------------------------------------------------------------------------
\section{Related Work}

In this section, we will briefly revisit the related topic, including denoising diffusion probabilistic models and single image generation.

\subsection{Denoising Diffusion Probabilistic Models}
As a new class of generative model, denoising diffusion probabilistic models~\cite{dhariwal2021diffusion} have achieved remarkable success on various tasks compared with generative adversarial nets~(GANs)~\cite{goodfellow2014generative}.
Diffusion model is a parameterized Markov chain that optimizes the lower variational bound on the likelihood function to generate samples matching real distribution.
Ho~\emph{et. al.}~\cite{ho2020denoising} first propose diffusion model and Dhariwal and Nichol~\cite{dhariwal2021diffusion} further show the potential of diffusion models, achieving better image sample quality compared with other generative models, \emph{i.e.}, GAN, on ImageNet dataset~\cite{deng2009imagenet}.
After that, more and more researchers turn their attention to diffusion models~\cite{nichol2021improved, song2020denoising, gu2021vector, saharia2021image, nichol2021glide, saharia2021palette, li2022srdiff, sasaki2021unit, kim2021diffusionclip, meng2021sdedit, ho2021classifier, lugmayr2022repaint, ho2022cascaded, kingma2021variational, kong2021fast, san2021noise, austin2021structured}.
Saharia~\emph{et. al.}~\cite{saharia2021image} achieve success in super-resolution with diffusion models.
Pattle~\cite{saharia2021palette} explores diffusion models on four image-to-image translation problems, \emph{i.e.}, colorization, inpainting, uncropping, and JPEG decompression.
And two concurrent works \cite{gu2021vector,nichol2021glide} apply diffusion models for text-to-image generation problem.

The above approaches deal with conditional image generation by directly training on corresponding datasets.
Diffusion models are also capable of solving conditional image generation problem by taking advantage of pre-trained unconditional models.
Bahjat~\emph{et. al.}~\cite{kawar2022denoising} propose an unsupervised posterior sampling method, \emph{i.e.}, DDRM, to solve any linear inverse problem, \emph{e.g.}, image inpainting and colorization, with a pre-trained diffusion model.
ILVR~\cite{choi2021ilvr} guides the generative process in a diffusion model to generate high-quality images based on a given reference image.
DDIBs~\cite{su2022dual} and EGSDE~\cite{zhao2022egsde} apply pre-trained diffusion models to unpaired image-to-image translation task. 
And~\cite{avrahami2021blended} introduces multi-modal information like text as the guidance of the generative process to generate text-related images.
With the characteristic of diffusion models, our SinDiffusion can also generate condition-related images to solve a variety of image manipulation tasks.

\subsection{Single Image Generation}
Single image generation~\cite{shocher2019ingan, hinz2021improved, zhang2021exsingan, granot2022drop, zhang2022petsgan, lin2020tuigan, asano2019critical, yoo2021sinir, chen2021mogan, sun2020esingan, zheng2021patchwise, sushko2021generating} aims to generate diverse results by learning the internal patch distribution from a single image. 
The groundbreaking work on this problem is SinGAN~\cite{shaham2019singan} which first explores this problem and proposes a series of PatchGAN~\cite{isola2017image} based on an image pyramid to generate diverse results hierarchically.
A concurrent work, \emph{i.e.}, InGAN~\cite{shocher2019ingan}, train a conditional GAN to solve the same problem based on a geometry transformation.
After that, more and more researchers turn their attention to this emerging topic.
ExSinGAN~\cite{zhang2021exsingan} train three modular GANs to model the distributions about structure, semantics and texture for learning an explainable generative model.
ConSinGAN~\cite{hinz2021improved} improves SinGAN by concurrently training several stages in a sequential multi-stage manner.
However, these methods are based on multi-scale GAN structure in SinGAN, which may accumulate errors and lead to unsatisfactory generation results with characteristic artifacts.
To this end, we propose a framework based on diffusion model to generate photorealistc and diverse results from a single natural image.

%-------------------------------------------------------------------------
\section{Methodology}
In this paper, we present a novel framework named SinDiffusion to learn internal distribution from a single natural image.
Unlike the progressive growing design in prior work, SinDiffusion is trained with a single denoising model at a single scale, which prevents the accumulation of errors.
Furthermore, we identify that a patch-level receptive field of the diffusion network plays an important role in capturing internal patch distribution and design a new denoising network structure.
Based on these two core designs, SinDiffusion generates high-quality and diverse images from a single natural image.
The rest of this section is organized as follows: 
We begin with revisiting SinGAN and presenting the motivation of SinDiffusion. 
After that, we introduce the structure design of SinDiffusion.

\begin{figure}[t]
  \centering
   \includegraphics[width=\linewidth]{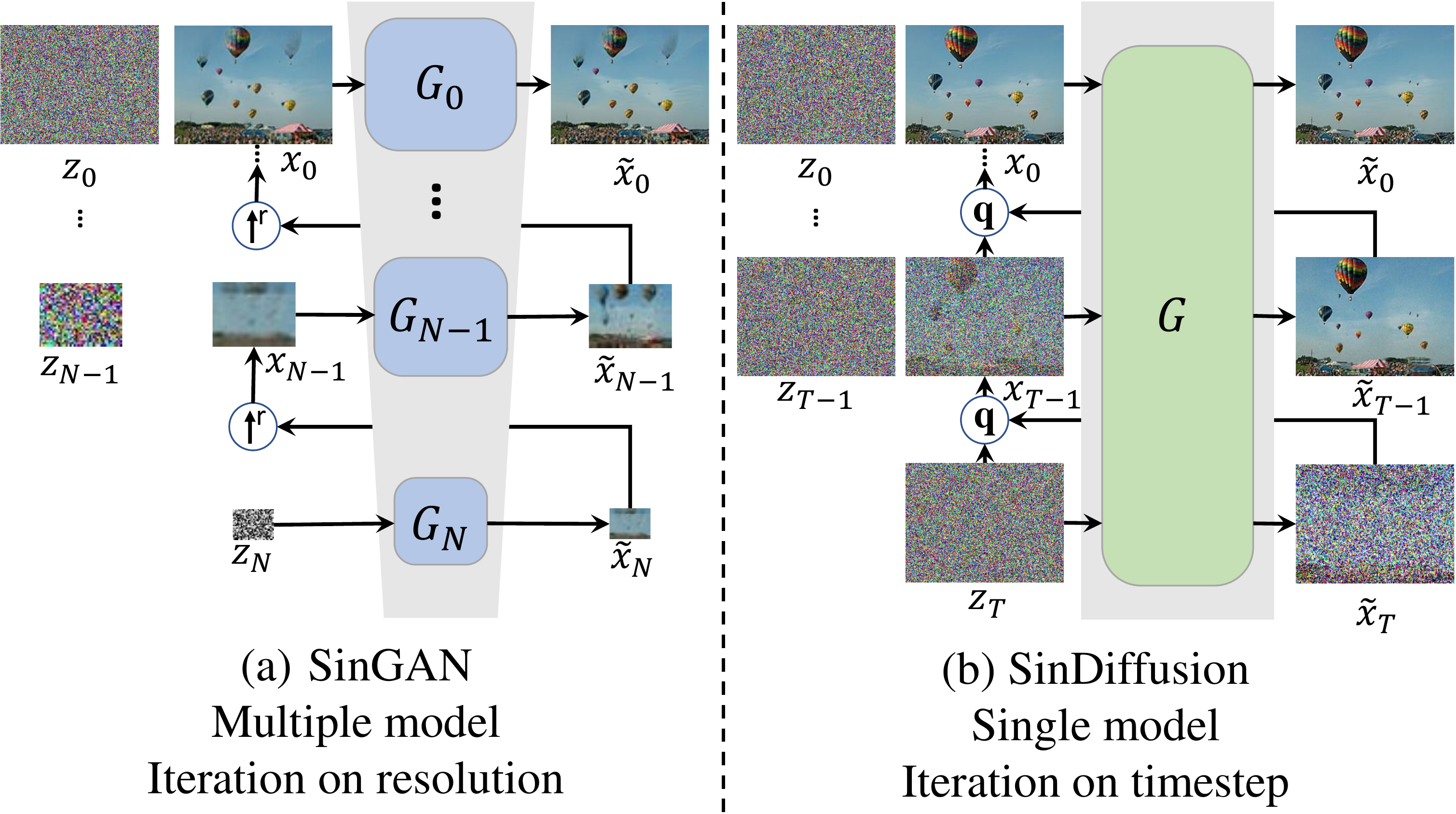}
   \vspace{-6mm}
   \caption{
   Comparison between SinGAN and SinDiffusion.
   $\uparrow^r$ refers to the upsample operation and $q$ refers to $q(x_{t-1} | x_t, x_0)$ in diffusion model.
   SinGAN generates the images hierarchically on the image resolution, which can also be seen as a kind of diffusion model.
   Inspired by this, we utilize diffusion model which generates images progressively on the timestep.
   }
   \vspace{-5mm}
   \label{fig:framework}
\end{figure}

\subsection{Revisiting SinGAN}
We begin with briefly revisiting SinGAN.
Figure~\ref{fig:framework}(a) presents the generation procedure of SinGAN.
To generate diverse images from a single image, one key design of SinGAN is to build an image pyramid and progressively grow the resolution of the generated image.
At each scale $n$, the image from the previous scale $\widetilde{x}_{n+1}$ is upsampled and fed into PatchGAN along with the input noise map $z_n$ to generate the image at scale $n$, which is formulated as follow,
\begin{equation}
    \widetilde{x}_{n} = G_{N}(\alpha_n(\widetilde{x}_{n+1})\uparrow^r + (1 - \alpha_n)z_n),
\end{equation}
where $\alpha_n$ is a blending factor that decreases as $n$ decreases.

Suppose there are small detail errors in the output image at a small scale. With the output resolution growing, these detail errors will be accumulated in the output image, thus causing an unsatisfactory final output. Through this analysis, we find the recently developed denoising diffusion probabilistic models (DDPMs)~\cite{ho2020denoising} do not suffer from this issue since they naturally process detail errors during the diffusion process.

%we propose a single-scale training framework for learning from a single natural image.

To this end, we propose a novel framework named SinDiffusion in Figure~\ref{fig:framework}(b).
Different from SinGAN, SinDiffusion performs the multiple-step generation process with a single denoising network at a single scale.
Although SinDiffusion also adopts the multiple-step generation process like SinGAN, the generated results are high-quality. This is because the diffusion model is based on a systematic derivation of mathematical equations, and errors arising from intermediate steps are repeatedly refined as noise during the diffusion process.

\begin{figure}[t]
  \centering
   \includegraphics[width=\linewidth]{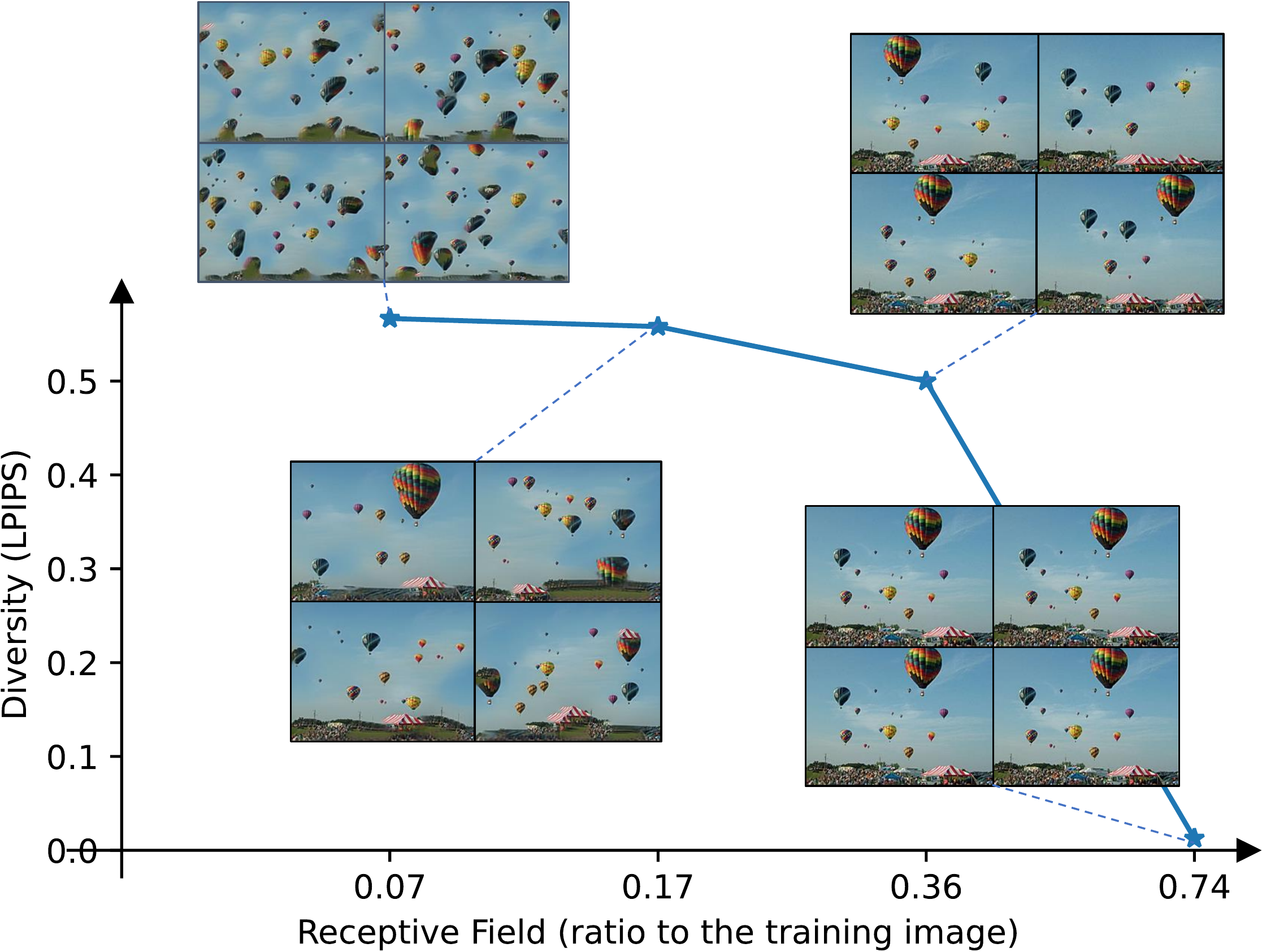}
   \vspace{-4mm}
   \caption{
   Relationship between the generation diversity and the receptive field of denoising network.
   With a smaller receptive field, SinDiffusion generates more diverse generation results.
   However, with very minor receptive field, SinDiffusion may learn very detailed patch distribution, which leads to the generated images not preserving the internal structure of the training image.
   }
   \vspace{-4mm}
   \label{fig:receptive}
\end{figure}

\subsection{SinDiffusion}
\label{sec:sindiffusion}

% To generate diverse images from a single natural image with a diffusion model, the receptive field of the denoising network plays an important role.
SinDiffusion is a diffusion-based model trained on a single scale to learn the internal patch statistics.
We first revisit the commonly-used diffusion model~\cite{dhariwal2021diffusion} and find that they tend to generate exactly the same images as the training image.
This is because the denoising network possesses strong capability with a large receptive field that covers the full image, which leads to the network memorizing the training image instead of learning the internal patch distribution.
Motivated by this, we suppose that the receptive field plays an important role in learning the internal patch statistics, and a patch-level receptive field of the diffusion network is crucial and effective for single image generation.

We study the relationship between generation diversity and the receptive field of the denoising network.
The receptive field is varied by modifying the network structure of the denoising network.
We design four network structures with different receptive fields but comparable capabilities and train these models on a single natural image.
Figure~\ref{fig:receptive} shows the results generated by the model with different receptive fields.
It is observed that, with a smaller receptive field, SinDiffusion tends to generate more diverse generation results and vice versa.
However, we find that the model of an extremely small receptive field can not preserve the reasonable structure of the image.
Therefore, a suitable receptive field is important and necessary to capture reasonable patch statistics.

\begin{figure}[t]
  \centering
   \includegraphics[width=\linewidth]{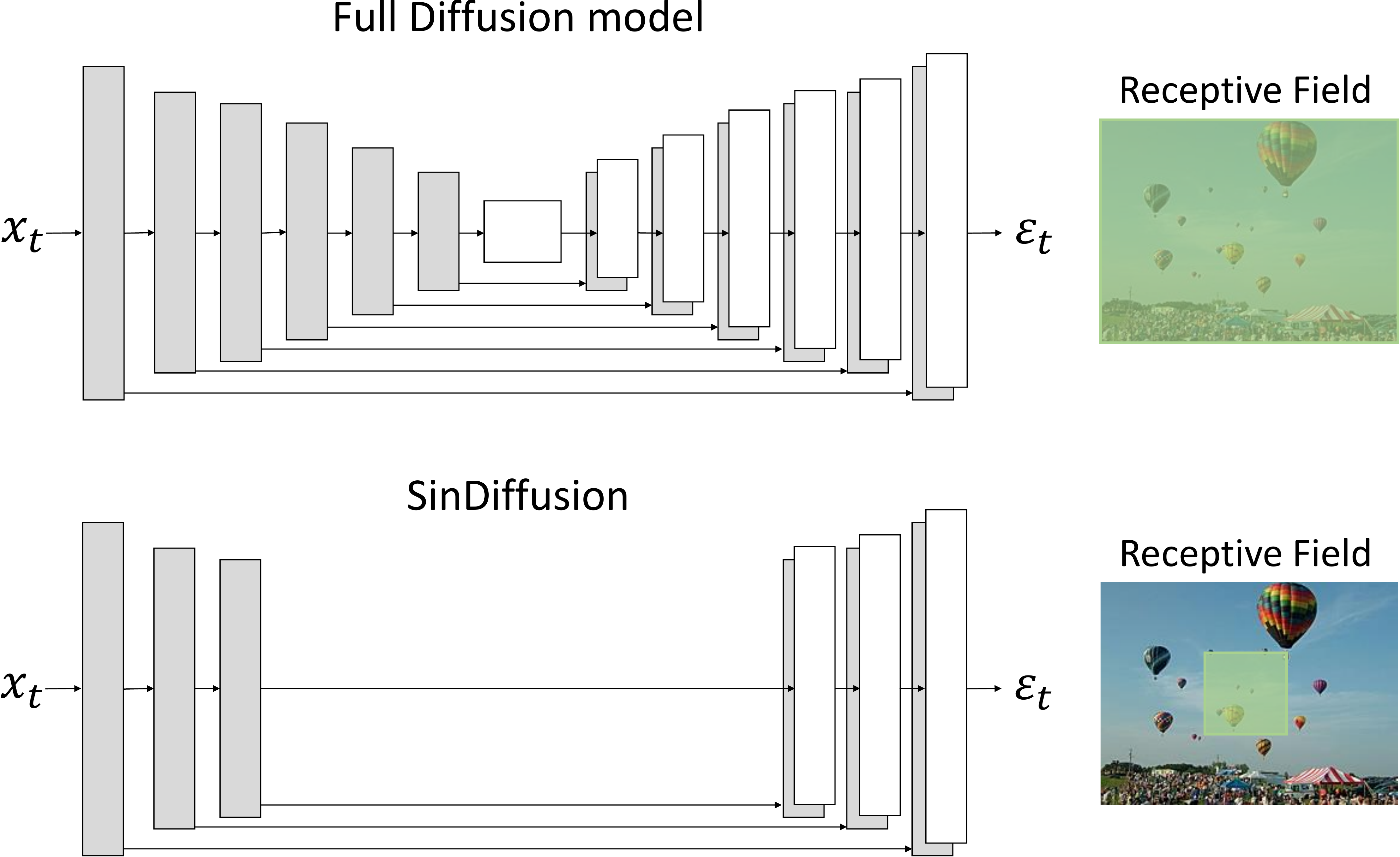}
   \vspace{-6mm}
   \caption{
   Comparison between vanilla diffusion model and SinDiffusion.
   The receptive field of a vanilla diffusion model covers the whole training image, which will lead to generating the same images as the training image.
   SinDiffusion learns to estimate the noise based on a local patch and finally generates diverse images.
   }
   \vspace{-4mm}
   \label{fig:framework2}
\end{figure}

Based on the above analysis, we redesign the commonly-used diffusion model and introduce a patch-wise denoising network for single image generation.
Figure~\ref{fig:framework2} presents an overview of patch-wise denoising network in SinDiffusion and the main difference compared with the previous denoising network.
First, we reduce the depth of the denoising network by lessening the downsample and upsample operation, which greatly expands the receptive field.
Meanwhile, the attention layer originally used in the deep layers of the denoising network is naturally removed, which makes SinDiffusion a fully convolutional network applicable to the generation of arbitrary resolution.
Second, we further limit the receptive field of SinDiffusion by reducing the time-embedded resblocks in each resolution.
In this way, we draw a patch-wise denoising network with a proper receptive field, generating photorealistic and diverse results.

\begin{figure*}[t]
  \centering
   \includegraphics[width=\linewidth]{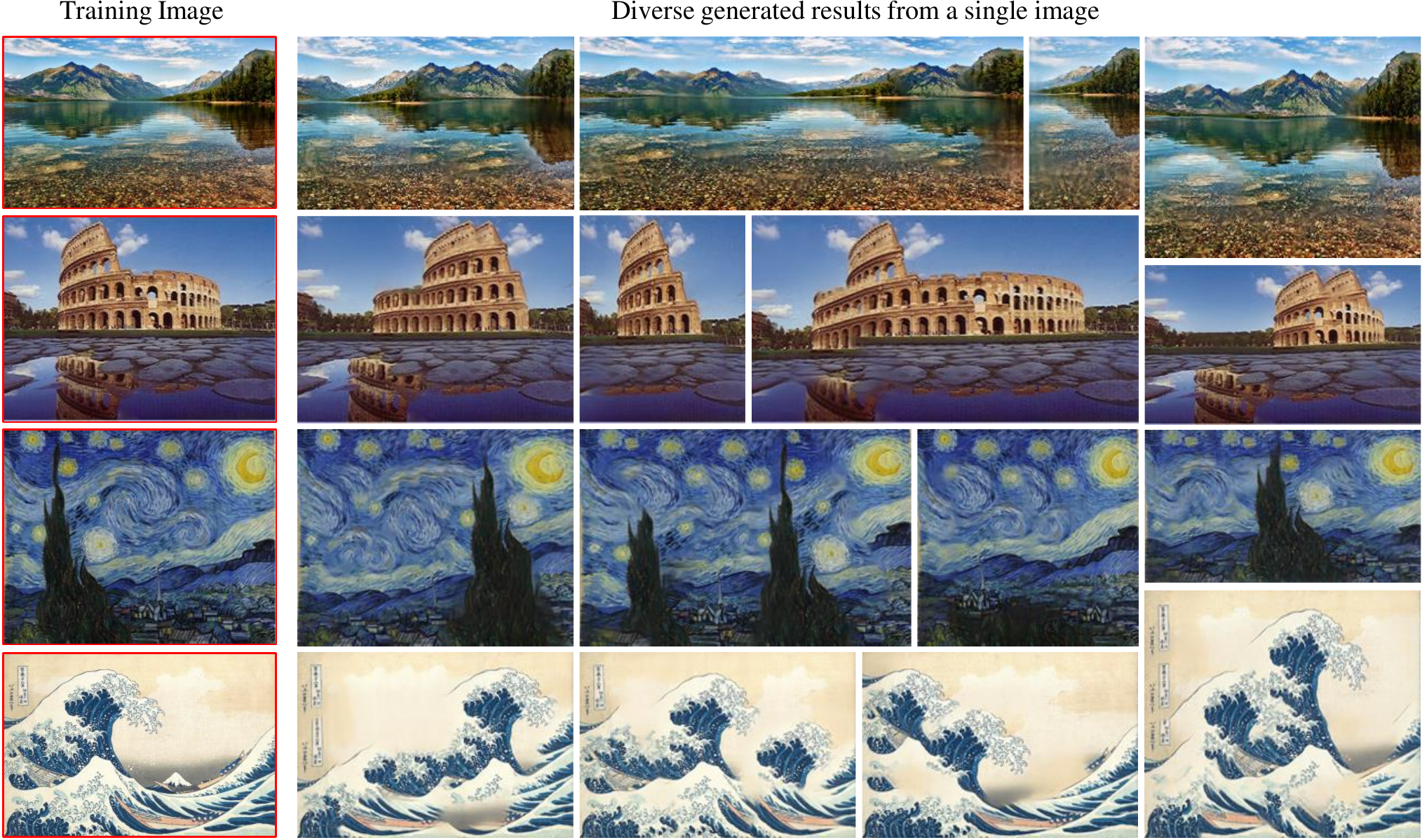}
   \vspace{-7mm}
   \caption{
   \textbf{Random samples from single images.}
   We perform experiments on the image of natural images and famous arts.
   It is observed that, for different resolutions, SinDiffusion generates diverse and realistic images which have similar patches to the training image. 
   }
   \vspace{-4mm}
   \label{fig:diverse}
\end{figure*}

\begin{figure*}[t]
  \centering
   \includegraphics[width=\linewidth]{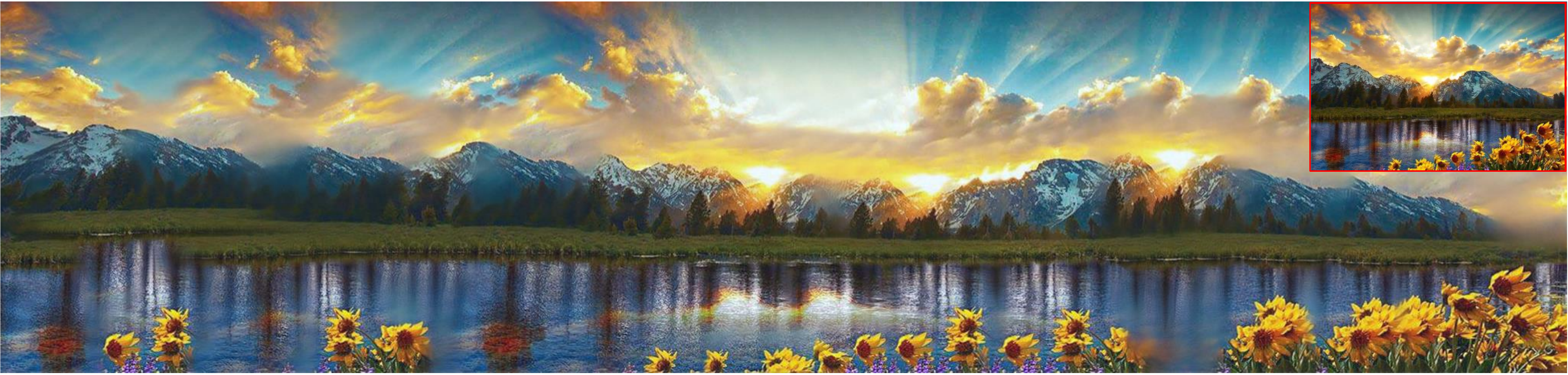}
   \vspace{-7mm}
   \caption{
   \textbf{High-resolution single image generation.}
   The resolution of the training image is $486 \times 741$.
   We utilize a SinDiffusion with higher network capacity and larger receptive field to accommodate high-resolution image generation.
   It is observed that our framework generates a realistic high-resolution image ($486 \times 2048$) that contains similar patterns and structure to the training image.
   }
   \vspace{-5mm}
   \label{fig:highres}
\end{figure*}

We train our SinDiffusion with the original denoising loss.
In diffusion models, given a training image $x$ and a random timestep $t \in \{0, 1, \dots, T\}$, a noisy version of the image $\widetilde{x}$ is produced as follows,
\begin{equation}
    \widetilde{x} = \sqrt{\alpha_t} x + \sqrt{1 - \alpha_t} \epsilon,
\end{equation}
where $\epsilon$ is a noise sampled from the standard Gaussian distribution.
$\alpha_t$ is a noise scheduler at timestep $t$.
$T$ is set to 1000 in our SinDiffusion.
SinDiffusion is trained to reconstruct the training image $x$ by predicting the involved noise $\epsilon$ with the timestep $t$, which is formulated as follows,
\begin{equation}
    \mathcal{L} = \mathbb{E}_{t, \epsilon}[\|\epsilon - \epsilon_\theta(\widetilde{x}, t)\|_2].
\end{equation}
Once trained, SinDiffusion can generate diverse images by an iterative denoising process, which is formulated as follows,
\begin{equation}
    x_{t-1} = \frac{1}{\sqrt{\alpha_t}}(x_t - \frac{1 - \alpha_t}{\sqrt{1-\beta_t}}\epsilon_\theta(x_t) + \mathbf{z}_t),
\end{equation}
where $\alpha_t$ and $\beta_t$ is the variance schedule factor in diffusion model.
$\mathbf{z}_t$ denotes a Gaussian noise involved at timestep $t$.

\begin{figure*}[t]
  \centering
   \includegraphics[width=\linewidth]{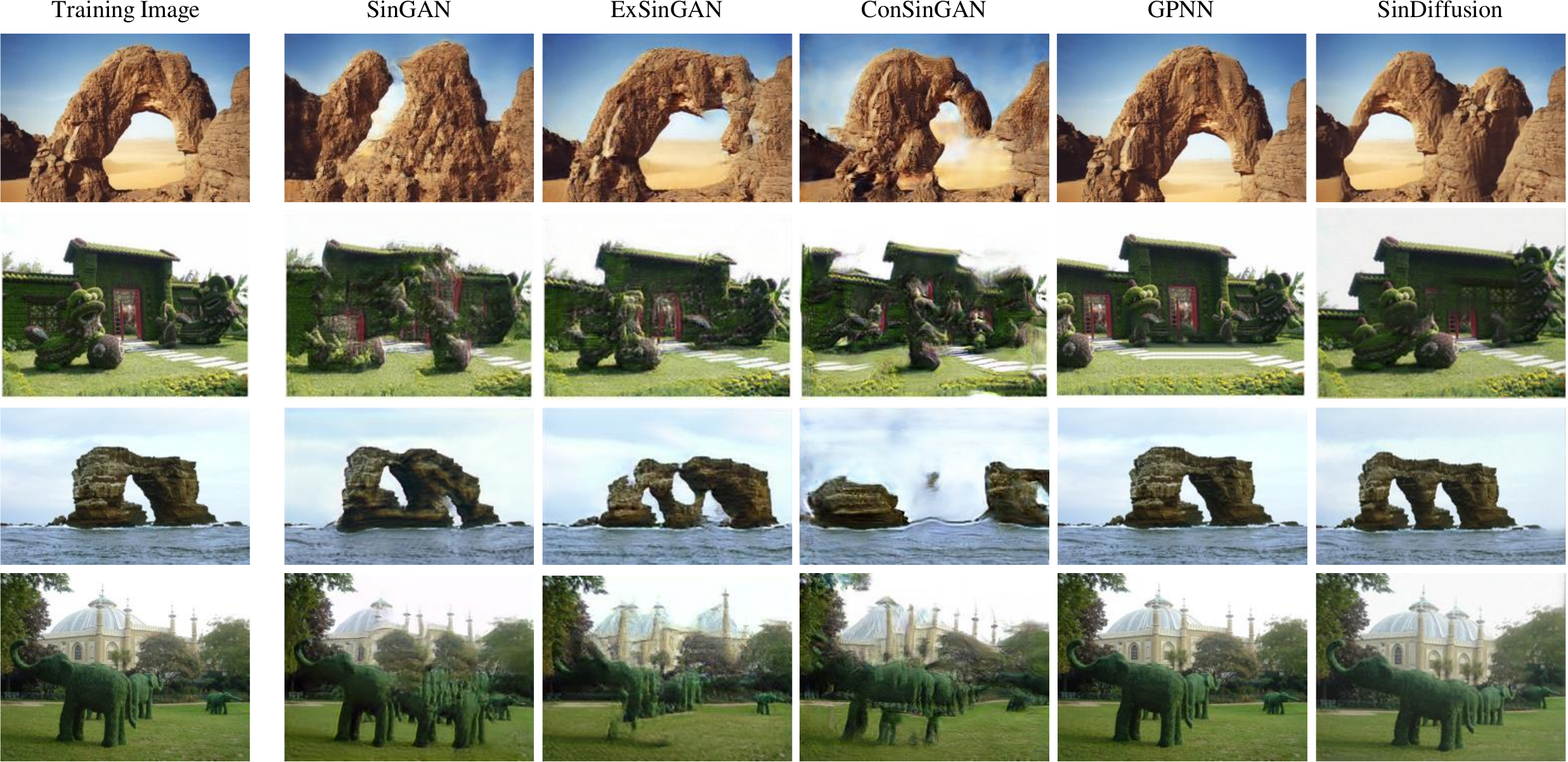}
   \vspace{-7mm}
   \caption{
   \textbf{Qualitative comparison on Places50 dataset.}
   We compare SinDiffusion with several challenging methods, \emph{i.e.}, SinGAN~\cite{shaham2019singan}, ExSinGAN~\cite{zhang2021exsingan}, ConSinGAN~\cite{hinz2021improved} and GPNN~\cite{granot2022drop}.
   SinGAN, ExSinGAN and ConSinGAN generate results with artifact due to error accumulation in multi-scale structure.
   GPNN is short of generalization and produces images close to the training images.
   By comparison, our generated images show superior performance on both fidelity and diversity.
   }
   \vspace{-6mm}
   \label{fig:qualitative}
\end{figure*}

%------------------------------------------------------------------------
\section{Experiments}
\label{sec:formatting}
\subsection{Setup}
\noindent \textbf{Datasets.} 
To evaluate the effectiveness of our method, we conduct experiments on various natural images collected online, including landscapes and arts.
Additionally, to systematically evaluate the quantitative performance, we also experiment on a dataset of natural landscapes, \emph{i.e.}, Places50.
Places50 is the set of 50 landscapes image used in SinGAN (50 images from Places365 dataset~\cite{zhou2018places}).
We train each SinDiffusion model on each image in Places50 and evaluate the fidelity and diversity of generated results.

\noindent \textbf{Implementation details.}
We train our SinDiffusion with AdamW optimizer~\cite{IlyaLoshchilov2018DecoupledWD}.
During training, we adopt an exponential moving average (EMA) with 0.9999 decay.
The whole framework is implemented by Pytorch and the experiments are performed on NVIDIA Tesla V100.

\noindent \textbf{Evaluation metric.}
We aim to assess both visual quality and diversity of generated images.
For the visual quality, following SinGAN~\cite{shaham2019singan}, we adopt the single-image Frechet Inception Distance (SIFID) metric.
Similar to FID, SIFID measures the deviation between the distribution of patch-wise features from the generated images and the real images.
To evaluate the generation diversity, we compute the average distance measured by the LPIPS metrics~\cite{zhang2018unreasonable} between multimodal generation results.
%------------------------------------------------------------------------

\begin{table}[t]
    \footnotesize
    \centering
    \begin{tabular}{l @{\hskip 17mm} c @{\hskip 17mm} c}
    \toprule
    \textbf{Method} & \textbf{SIFID} $\downarrow$ & \textbf{LPIPS} $\uparrow$ \\
    \midrule
    {SinGAN~\cite{shaham2019singan}} & 0.09 & 0.266 \\
    {ExSinGAN~\cite{zhang2021exsingan}} & 0.10 & 0.248 \\
    {ConSinGAN~\cite{hinz2021improved}} & \textbf{0.06} & 0.305 \\
    {GPNN~\cite{granot2022drop}} & 0.07 & 0.256 \\
    {SinDiffusion~(Ours)} & \textbf{0.06} & \textbf{0.387} \\
    \bottomrule
    \end{tabular}
    \vspace{-3mm}
    \caption{\textbf{Quantitative comparison with existing methods on single image generation.}
    $\uparrow$ indicates the higher the better, while $\downarrow$ indicates the lower the better.
    Notably, our method achieves state-of-the-art performance on SIFID and LPIPS. 
    }
    \vspace{-6mm}
    \label{tab:quantitative}
\end{table}

\subsection{Qualitative Evaluation}
Qualitative results of random generated images from SinDiffusion are shown in Figure~\ref{fig:diverse}, and more qualitative results are included in Supplementary Material.
We train our SinDiffusion on the image of natural images and famous arts.
For each training image, we first present a generated image under the same aspect ratio.
Then, we generate images of different aspect ratios with the training image to demonstrate the generalization of our SinDiffusion at different resolutions.
It is observed that, for different resolutions, our SinDiffusion can generate realistic images which have similar patterns to the training image. 

Furthermore, we explore SinDiffusion for generating high-resolution images from a single image.
Figure~\ref{fig:highres} presents the training image and generated result.
The training image is a landscape image of $486 \times 741$ resolution which contains rich components, \emph{i.e.}, clouds, mountains, grass, flowers, and a lake.
To accommodate the high-resolution image generation, we extend the SinDiffusion to an enhanced version, which has larger receptive fields and network capability.
Compared with the structure in Figure~\ref{fig:framework}, the enhanced version has 4 downsample layers and an additional time-embedded resblock on each scale.
With the enhanced SinDiffusion, we generate a high-resolution long-scroll image of $486 \times 2048$ resolution.
From Figure~\ref{fig:highres}, it is observed that our result maintains the internal layout of the training image and generalize new content.

\begin{table}[t]
    \footnotesize
    \centering
    \begin{tabular}{l @{\hskip 30mm} c}
    \toprule
    \textbf{Method} & \textbf{Preference} \\
    \midrule
    {SinDiffusion v.s. SinGAN} & 65.0\% \\
    {SinDiffusion v.s. ExSinGAN} & 77.5\% \\
    {SinDiffusion v.s. ConSinGAN} & 81.0\% \\
    {SinDiffusion v.s. GPNN} & 73.0\% \\
    \bottomrule
    \end{tabular}
    \vspace{-3mm}
    \caption{\textbf{Paired user study on the Places50 dataset.}
    It is observed that our method is clearly preferred over the competitors.}
    \vspace{-6mm}
    \label{tab:userstudy}
\end{table}

\begin{figure*}[t]
  \centering
   \includegraphics[width=\linewidth]{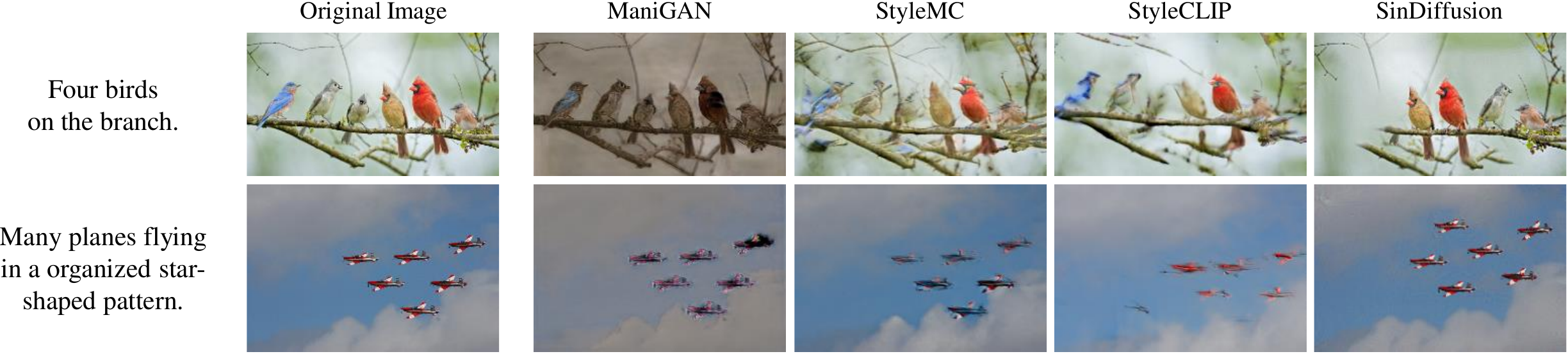}
   \vspace{-7mm}
   \caption{
   \textbf{Qualitative comparison on text-guided image generation.}
   We compare SinDiffusion with several previous methods, \emph{i.e.}, ManiGAN~\cite{li2020manigan}, StyleMC~\cite{kocasari2022stylemc} and StyleCLIP~\cite{patashnik2021styleclip}.
   It demonstrates that SinDiffusion can generate images which more corresponds with the input text compared with previous methods.
   }
   \vspace{-5mm}
   \label{fig:text-guided}
\end{figure*}

\begin{figure*}[t]
  \centering
   \includegraphics[width=\linewidth]{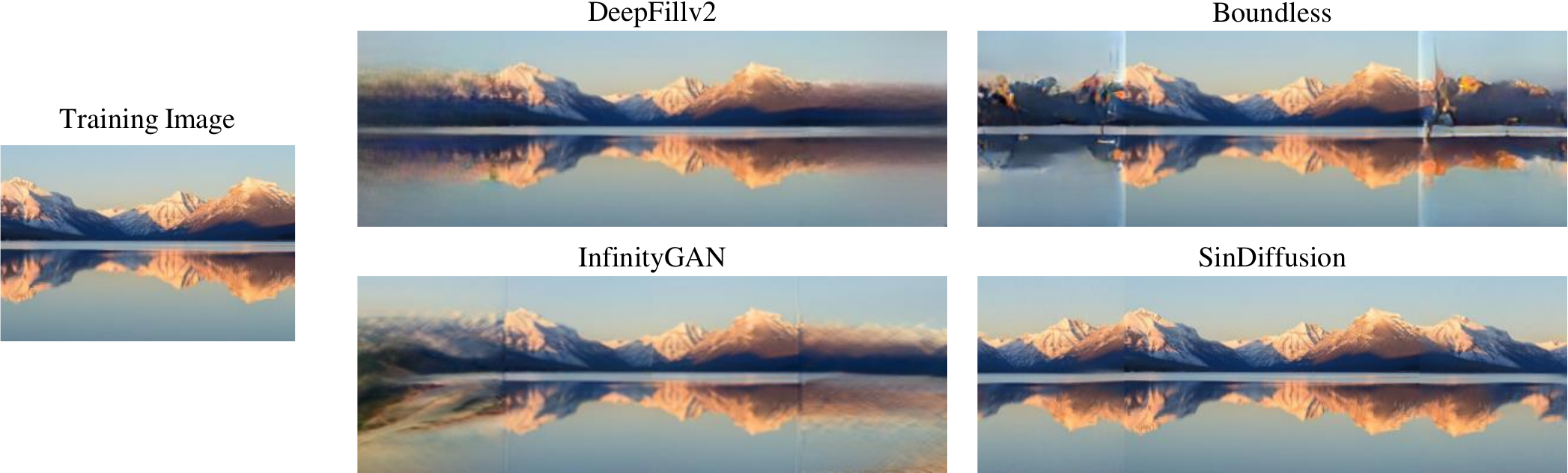}
   \vspace{-7mm}
   \caption{
   \textbf{Qualitative comparison on image outpainting.}
   We compare SinDiffusion with several previous methods, \emph{i.e.}, DeepFillv2~\cite{yu2019free}, Boundless~\cite{teterwak2019boundless} and InfinityGAN~\cite{lin2021infinitygan}.
   SinDiffusion can better predict the content outside the image compared with previous methods.
   }
   \vspace{-6mm}
   \label{fig:outpaint}
\end{figure*}

\subsection{Comparison with previous methods}
We compare our SinDiffusion with several challenging methods, \emph{i.e.}, SinGAN~\cite{shaham2019singan}, ExSinGAN~\cite{zhang2021exsingan}, ConSinGAN~\cite{hinz2021improved} and GPNN~\cite{granot2022drop}.
The quantitative results in shown in Table~\ref{tab:quantitative}.
With the help of progressive refinement, SinDiffusion achieves state-of-the-art performance compared with previous GAN-based method.
Notably, our method highly improves the diversity of generated images, surpassing the most challenging method by +0.082 LPIPS score on the average of 50 models trained on the Places50 dataset.

Besides the quantitative results, we also present the qualitative results on the Places50 dataset in Figure~\ref{fig:qualitative}.
The images generated by SinGAN, ExSinGAN and ConSinGAN show unreasonable structure and artifact in the details.
This is because of the error accumulation in the multi-scale structure, enlarging the artifact produced at the initial several scales.
GPNN is able to produce realistic images.
However, since GPNN clones the nearest patches from the training image, its generated images lose patch-level diversity and tend to be similar to the training images.
Unlike these methods, the images generated by SinDiffusion are with reasonable structures and sharp details and are capable of generalizing novel patterns from the training images.

Furthermore, we conduct a user study to evaluate the visual performance of generated images.
There are 20 volunteers participating in this study.
In the study, we present each volunteer with 10 pairs of generated results for each paired user study (40 pairs in total).
Volunteers are asked to answer this question, \emph{i.e.}, which group of images shows more diversity and better quality?
The voting results are reported in Table.~\ref{tab:userstudy}.
It is observed that our method is clearly preferred over competitors in more than 65\% of the time.

\subsection{Image Manipulation}
We explore the application of SinDiffusion on various image manipulation tasks.
We directly use our trained SinDiffusion model for all the applications, without architectural changes or further finetuning.
Different from SinGAN, injecting the condition image into the generation pyramid at some scale, SinDiffusion is utilized for various applications by designing the sampling procedure.
By virtue of this, besides the applications in SinGAN, \emph{i.e.}, image editing, image harmonization and image-to-image translation, SinDiffusion can be further applied to image manipulation tasks like text-guided image generation and image outpainting.
We will present the details as follow.

\noindent \textbf{Text-guided image generation.}
To generate the images from a single image corresponding to the given text, we guide the sampling procedure by a gradient from a pre-trained visual-linguistic model~$C(\cdot, \cdot)$, \emph{i.e.}, CLIP. 
Supposing a pre-trained diffusion model with estimated mean $\mu_\theta(x_{t-1} | x_t)$, an image that corresponds to a given text $L$ can be generated by perturbing the mean, which is formulated as follows,
\begin{equation}
    \hat{\mu}_\theta(x_{t-1}|x_t) = \mu_\theta(x_{t-1}|x_t) + s \cdot \nabla_{x_t} \text{log}~C(x_t, L),
\end{equation}
where the hyperparameter $s$ is the guidance scale, which balances the fidelity and correspondence with the given text. 

Figure~\ref{fig:text-guided} presents the text-guided image generation results of SinDiffusion and previous methods.
We train SinDiffusion on various images and use text as the condition to generate images with a different number of objects or different shapes from a single image.
From the figure, it is observed that, by changing the conditional text, SinDiffusion is able to controllably generate realistic images from the training image. 
By comparison, previous methods fail to generate corresponding images with text under the single-image setting.
This demonstrates that we provide an effective approach to control the single-image model through high-level semantics.

\noindent \textbf{Image outpainting.}
Image outpainting aims to generate content which resides beyond the edges of an image.
With iterative image outpainting, ideally, we can extend a finite-sized image to an infinite size.
Since our SinDiffusion model learns the internal distribution of the patches from the training image, it is inherently capable of imagining the content outside the given image.
Supposing a pre-trained diffusion model with iterative latent $x_\theta(z_t)$, we outpaint a natural image $x^a$ by replacing the given region, which is formulated as follows,
\begin{equation}
    \hat{x}_\theta(z_t) = x_\theta(z_t) \cdot m^a + x^a_{t-1} \cdot (1 - m^a),
\end{equation}
where $m^a$ indicates the outpainting region.
$x^a_{t-1}$ refers to the noisy version of the natural image $x^a$ at timestep $t-1$.

In Figure~\ref{fig:outpaint}, we compare SinDiffusion with some previous image outpainting methods, \emph{i.e.}, DeepFillv2~\cite{yu2019free}, Boundless~\cite{teterwak2019boundless} and InfinityGAN~\cite{lin2021infinitygan}.
From the figure, it is observed that, by learning the patch distribution, SinDiffusion generates reasonable and realistic images with content that conforms to the intrinsic distribution. 
In contrast, previous methods produce unrealistic and blurry results, failing to predict what is outside the original image.
This indicates that SinDiffusion is more efficient and flexible for image outpainting of a single natural image.

\noindent \textbf{Other image manipulation task.}
SinDiffusion can also be applied to image manipulation tasks in prior methods, \emph{i.e.}, image editing, image harmonization and image-to-image translation.
Inspired by~\cite{choi2021ilvr}, we generate from a reference image $y$ by designing the sampling procedure as follows,
\begin{equation}
    \hat{x}_\theta(z_t) = \phi_N(y_{t-1}) + x_\theta(z_t) - \phi_N(x_\theta(z_t)),
\end{equation}
where $\phi_N(\cdot)$ refers to a linear low-pass filtering operation.
$y_{t-1}$ refers to the noisy version of the reference image $y$ at timestep $t-1$.
Some image manipulation results from SinDiffusion are shown in Figure~\ref{fig:teaser}, and more results are included in Supplementary Material.

\subsection{Ablation Study}
We conduct ablative experiments to evaluate the effectiveness of several important designs in SinDiffusion, \emph{i.e.}, whether to utilize the multi-scale structure and the receptive field of SinDiffusion. We perform experiments on a subset of the Places50 dataset.

\noindent \textbf{Multi-scale v.s. Single-scale.} 
Different from previous multi-scale approaches, we introduce a diffusion model to solve the problem on a single scale.
To verify the efficiency of single-scale design, we design a baseline variant as the comparison.
As an alternative, we convert the GAN in each scale of SinGAN to a diffusion model, and thereby present a multi-scale diffusion model on single image generation.
From Table~\ref{tab:multiscale}, it can be seen that SinDiffusion achieves superior performance to the multi-scale diffusion model.
Meanwhile, compared with the multi-scale diffusion model, SinDiffusion has much smaller network parameters and computational consumption, which also indicates the advantage of the single-scale design.

\begin{table}[t]
    \footnotesize
    \centering
    \begin{tabular}{c @{\hskip 14mm} c @{\hskip 14mm} c}
    \toprule
    \textbf{Settings} & \textbf{SIFID} $\downarrow$ & \textbf{LPIPS} $\uparrow$ \\
    \midrule
    {Multi-scale} & 0.74 & 0.349 \\
    {SinDiffusion (Single-scale)} & \textbf{0.11} & \textbf{0.432} \\
    \bottomrule
    \end{tabular}
    \vspace{-3mm}
    \caption{\textbf{Ablation Study on the single-scale design.}
    $\uparrow$ indicates the higher the better, while $\downarrow$ indicates the lower the better.
    It is observed that SinDiffusion achieve superior performance on SIFID and LPIPS compared with multi-scale baselines. 
    }
    \vspace{-3mm}
    \label{tab:multiscale}
\end{table}

\noindent \textbf{Receptive field.}
In Section~\ref{sec:sindiffusion}, we have shown some examples on how the receptive field affects the generated results.
We will further perform more complete analysis here.
The quantitative results are reported in Table~\ref{tab:receptive}.
From the table, it is observed that SinDiffusion generates more diverse images with a smaller receptive field.
However, the fidelity of the generated image (SIFID) increases as the receptive field increases.
Therefore, We take a suitable receptive field to trade off generation quality and diversity.

\begin{table}[t]
    \footnotesize
    \centering
    \begin{tabular}{c @{\hskip 4mm} c @{\hskip 14mm} c}
    \toprule
    \textbf{Receptive field (to the full image)} & \textbf{SIFID} $\downarrow$ & \textbf{LPIPS} $\uparrow$ \\
    \midrule
    {0.07} & 1.56 & \textbf{0.581} \\
    {0.17} & 0.59 & 0.579 \\
    {0.36} & 0.11 & 0.432 \\
    {0.74} & \textbf{0.08} & 0.032 \\
    \bottomrule
    \end{tabular}
    \vspace{-3mm}
    \caption{\textbf{Ablation Study on the receptive field of SinDiffusion.}
    $\uparrow$ indicates the higher the better, while $\downarrow$ indicates the lower the better.
    Under a larger receptive field, SinDiffusion produces images with higher fidelity (SIFID) and lower diversity (LPIPS). 
    }
    \vspace{-6mm}
    \label{tab:receptive}
\end{table}

\vspace{-1mm}
\section{Conclusion}
In this paper, we present the first attempt to explore the diffusion model on single image generation and propose a novel framework named Single-image Diffusion Model~(SinDiffusion).
In particular, we find that the receptive field plays an important role in diverse image generation and design a patch-wise denoising network for producing realistic and diverse images.
Furthermore, with the trained SinDiffusion model, we study a variety of image manipulation tasks, \emph{i.e.}, text-guided image generation, and image outpainting.
Extensive experiments on various natural images and the Places50 dataset demonstrate the effectiveness of our method.
Our method achieves state-of-the-art performance in terms of SIFID and LPIPS metrics and shows a better visual quality of generated images compared with previous methods.
The performance on image manipulation further demonstrates the usefulness and flexibility of SinDiffusion.

%%%%%%%%% REFERENCES
{\small
\bibliographystyle{ieee_fullname}
\bibliography{egbib}
}

\clearpage

This supplementary material provides more details not included in the main paper due to space limitations.
In the following, we first introduce the implementation details.
Then we will present more qualitative experiment results. 

\section*{A.~Implementation Details}
In this section, we provide more implementation details on SinDiffusion, including details on the denoising network and diffusion procedures.

\vspace{2mm}
\noindent \textbf{Denoising network.}
In Section~3.2, we introduce the patch-wise denoising network in SinDiffusion, which is a U-Net-based network estimating the noise in the input noisy image.
The encoder and decoder of the denoising network are 3 stages and the spatial resolution of the feature on each layer is 1, 1/2, and 1/4 of the input resolution, respectively.
The channel of each layer is set to 64, 128, and 256, respectively.
Each encoder layer contains 1 time-embedded Resblocks while each decoder layer contains 2 time-embedded Resblocks.
In high-resolution single image generation in Section~4.2, we utilize an enhanced denoising network, whose encoder and decoder are 4 layers.
The channel of each layer is set to 64, 128, 256, and 512, respectively.
Each encoder layer contains 2 time-embedded Resblocks while each decoder layer contains 3 time-embedded Resblocks.
In addition, we use half-precision float computation to accelerate training and reduce memory consumption.

\vspace{2mm}
\noindent \textbf{Diffusion procedure.}
As mentioned in Section~3, we propose a single-image diffusion model to learn the internal distribution from a single natural image.
Following DDPM [12], we set the total diffusion timestep $T$ to 1000.
In the forward process, the Gaussian noise is involved in the data according to a variance schedule $\beta_1, \dots, \beta_T$.
In our implementation, the variance schedule is arranged linearly with respect to the timestep $t$.
We adopt standard diffusion sampling in the diffusion process, introducing noise at each step and refining the generated image iteratively in 1000 steps.

\vspace{2mm}
\noindent \textbf{Evaluation metrics.}
As mentioned in Section~4.1, we apply SIFID and LPIPS metrics to evaluate the quality and diversity of generated images, respectively.
For SIFID metrics, we follow SinGAN [36] and use deep features at the output of the convolutional layer just before the second pooling layer in VGG19 network.
SIFID measures the distance between the patch distribution of two images in the feature space using a pre-trained Inception model, which is formulated as follows, 
\begin{equation}
    \mathrm{SIFID}(X, Y) = \| \mu_\mathbf{X} - \mu_\mathbf{Y} \|_2^2 + \mathrm{Tr}(\Sigma_\mathbf{X} + \Sigma_\mathbf{Y} - 2(\Sigma_\mathbf{X}\Sigma_\mathbf{Y})^{\frac{1}{2}}),
\end{equation}
where ($\mu_\mathbf{X}$, $\Sigma_\mathbf{X}$) and ($\mu_\mathbf{Y}$, $\Sigma_\mathbf{Y}$) refer to the mean value and covariance of patch distribution of the generated and training image, respectively.

For the LPIPS metric, we generate a set of images $x^N = \{x_i, i=1,2, \dots, N\}$ by multimodal generation. 
The diversity metric, \emph{i.e.}, LPIPS, is formulated as follows,
\begin{equation}
    \mathrm{LPIPS}(x^N) = \frac{2}{(N-1)(N-2)} \sum_{i=1}^N \sum_{j=i+1}^N \mathbf{d}(x_i, x_j),
\end{equation}
where $d(\cdot, \cdot)$ is a weighted perceptual similarity between two images, computed by the features extracted from a pre-trained AlexNet.
The number of images $N$ is set to 10 in our implementation.

\section*{B.~Additional Experiment Results}
In this section, we first present more qualitative results trained on single natural images.
Then, we supplement more single image manipulation task which is not shown in the main paper due to space limitations.

\subsection*{B.1~More Qualitative Results}
\begin{figure*}[t]
    \centering
    \includegraphics[width=\linewidth]{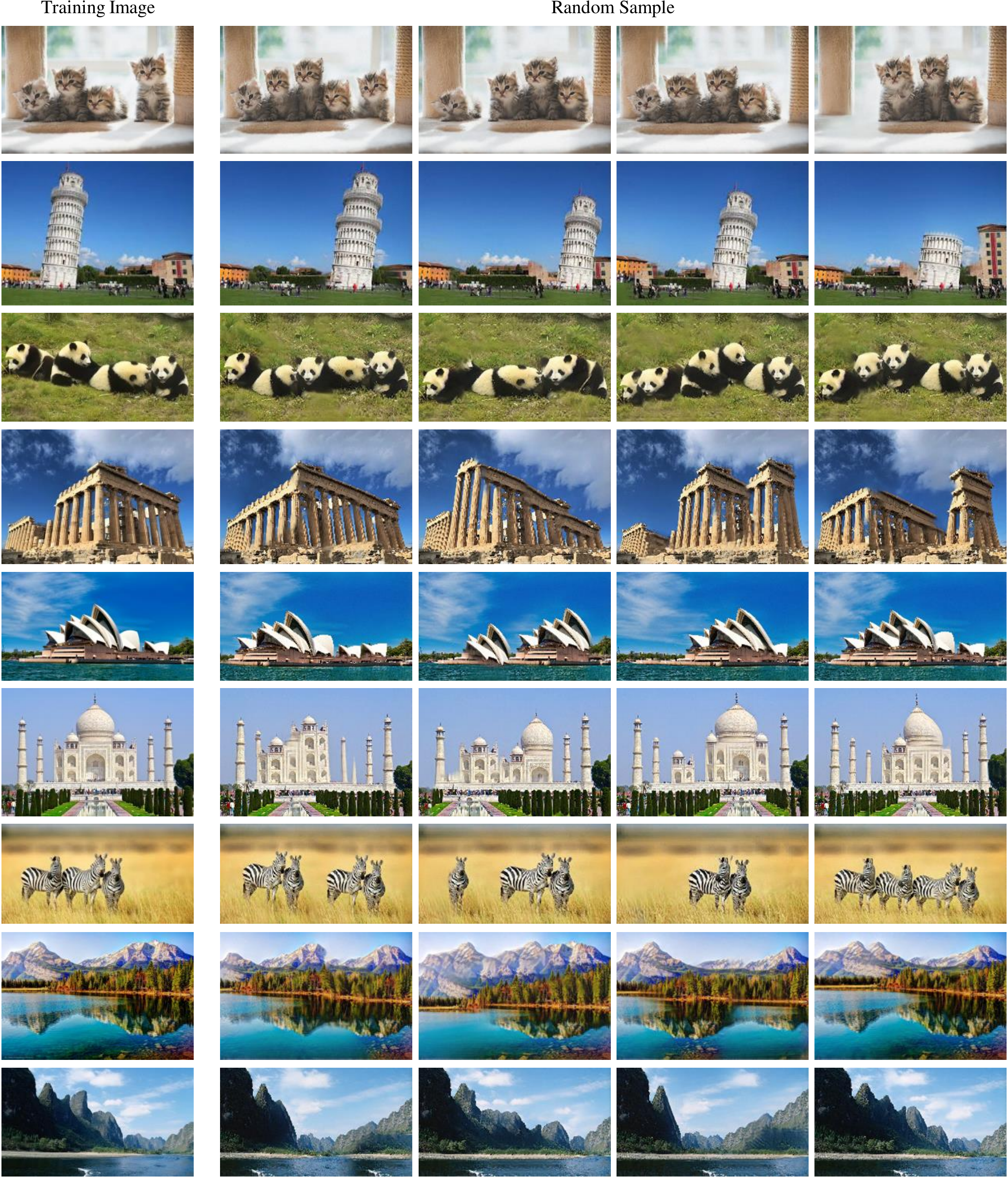}
    \vspace{-6mm}
    \caption{
    More randomly-sampled results of the same resolution as the training image.
    }
    \label{fig:sameres}
\end{figure*}

\begin{figure*}[t]
    \centering
    \includegraphics[width=\linewidth]{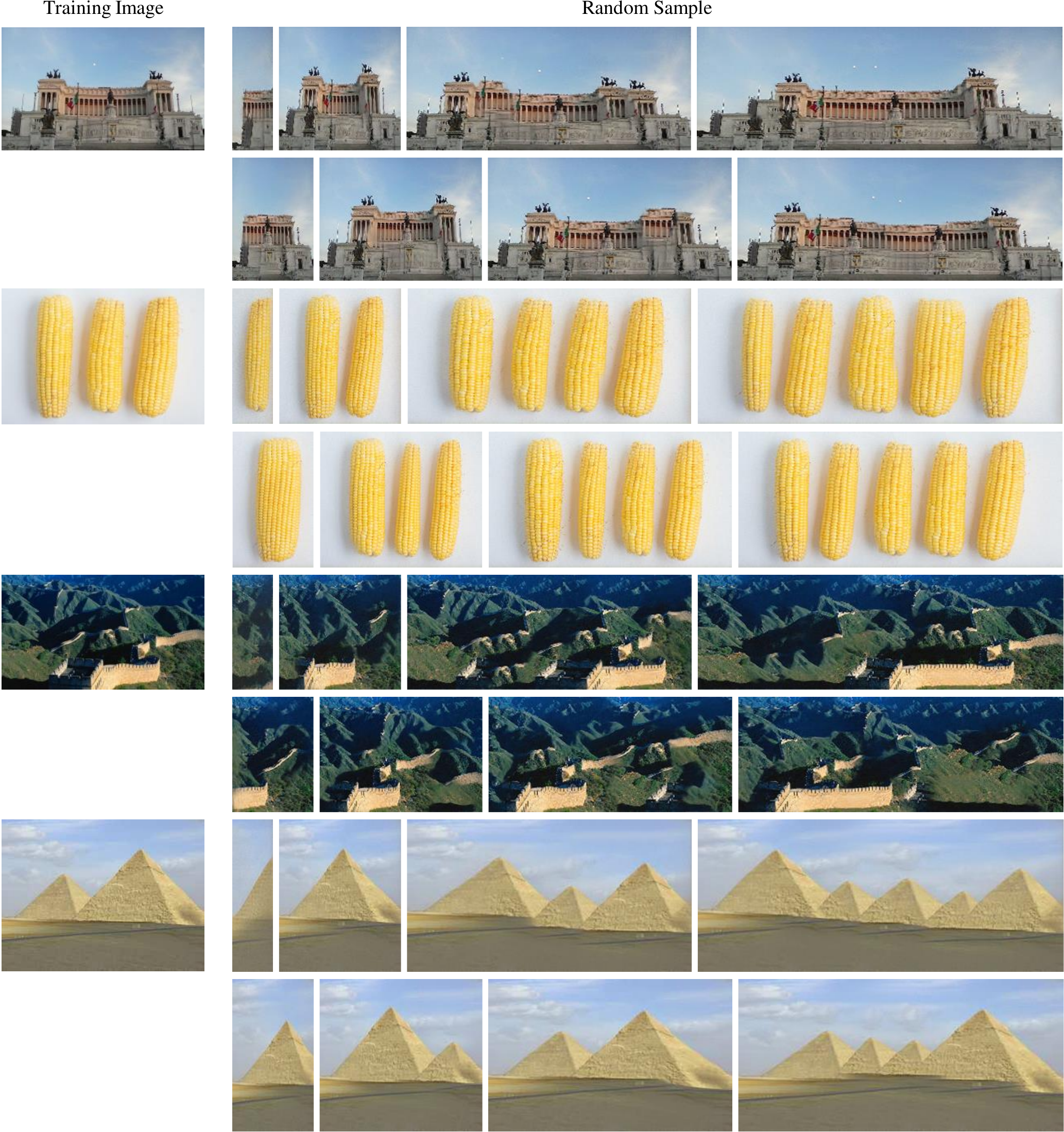}
    \vspace{-6mm}
    \caption{
    More randomly-sampled results of arbitrary resolutions with the training image.
    }
    \label{fig:abrares}
\end{figure*}

\begin{figure*}[t]
    \centering
    \includegraphics[width=\linewidth]{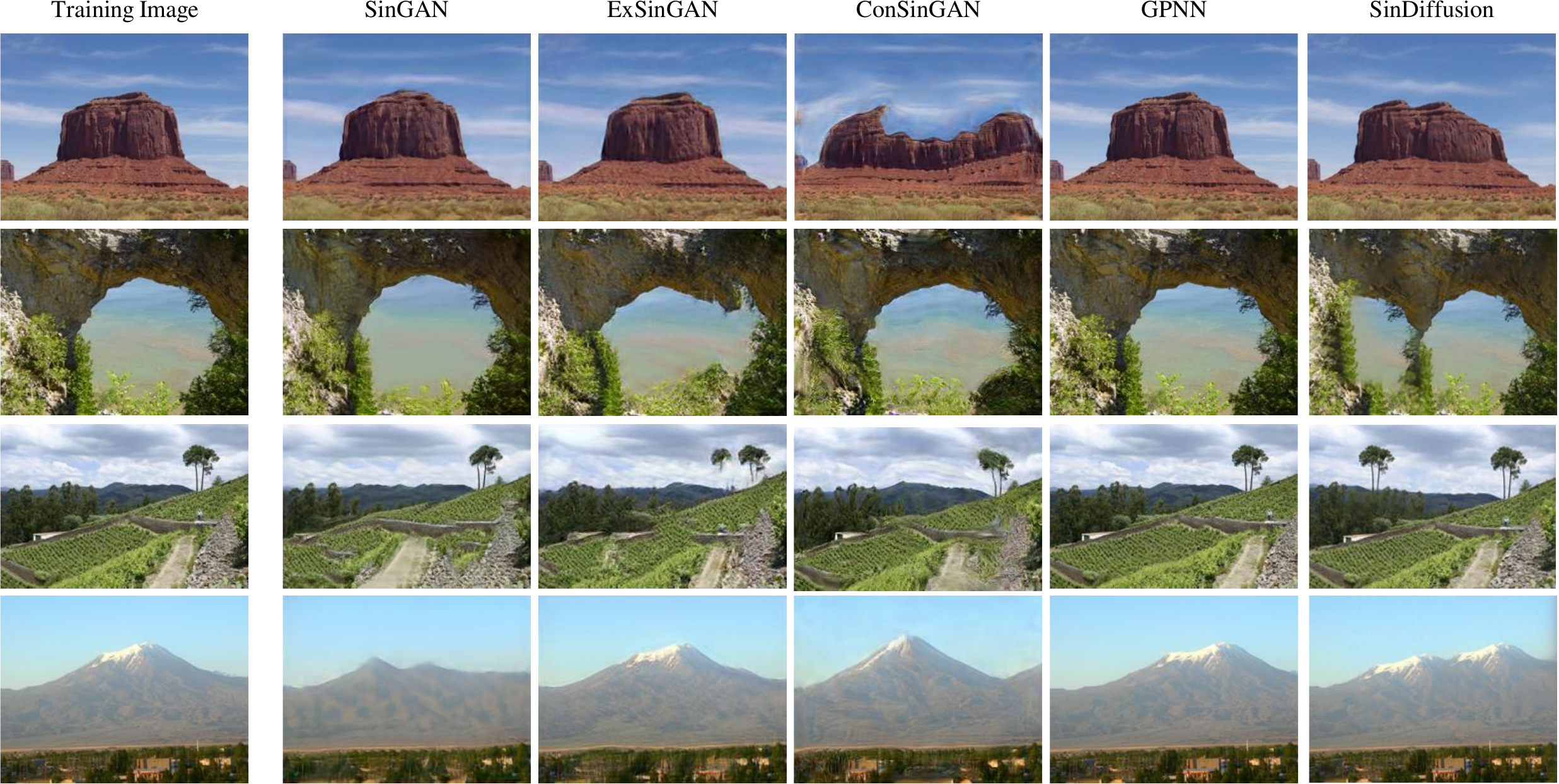}
    \vspace{-6mm}
    \caption{
    More comparison between SinDiffusion and previous methods on the Places50 dataset.
    SinDiffusion generates more realistic and diverse images compared with previous methods.
    Please zoom in for more details.
    }
    \label{fig:comp}
\end{figure*}

We present more qualitative results trained on single natural images with SinDiffusion.
Figure~\ref{fig:sameres} shows generated images under the same resolution as the training image.
Figure~\ref{fig:abrares} shows generated images of arbitrary resolutions.
Figure~\ref{fig:highres} shows high-resolution generated images.
Figure~\ref{fig:comp} shows more comparison results with previous methods, \emph{i.e.}, SinGAN, ExSinGAN, ConSinGAN, and GPNN.

\subsection*{B.2~Image Manipulation}
In this subsection, we present more cases on image manipulation task, \emph{i.e.}, text-guided image generation, image outpainting and paint-to-image translation.

\vspace{2mm}
\noindent \textbf{Text-guided image generation.}
As shown in Section 4.4, we explore text-guided image generation.
We train a SinDiffusion on a single natural image and generalize from the training image according to the given text.
We further show some text-guided image generation results in Figure~\ref{fig:text-guided}.

\begin{figure*}[t]
    \centering
    \includegraphics[width=\linewidth]{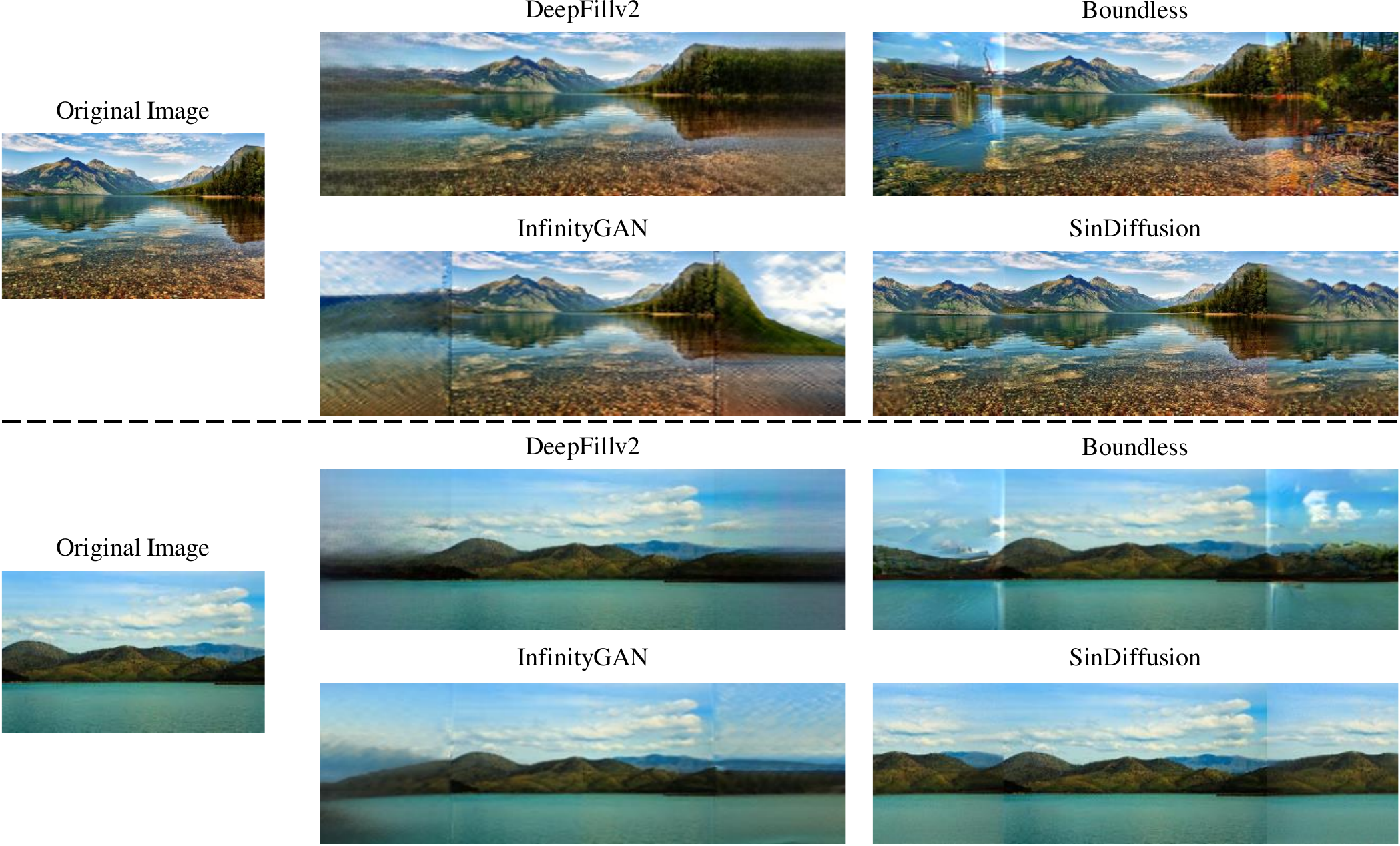}
    \vspace{-6mm}
    \caption{
    More comparison between SinDiffusion and previous methods on image outpainting.
    }
    \vspace{-3mm}
    \label{fig:outpainting}
\end{figure*}

\vspace{2mm}
\noindent \textbf{Image outpainting.}
As mentioned in Section 4.4, image outpainting aims to generate content beyond the edges of an image.
We train a SinDiffusion on a single natural image and generate what is outside the training image by replacing the given region during the sampling procedure.
We show more qualitative results in Figure~\ref{fig:outpainting}.

\begin{figure*}[t]
    \centering
    \includegraphics[width=\linewidth]{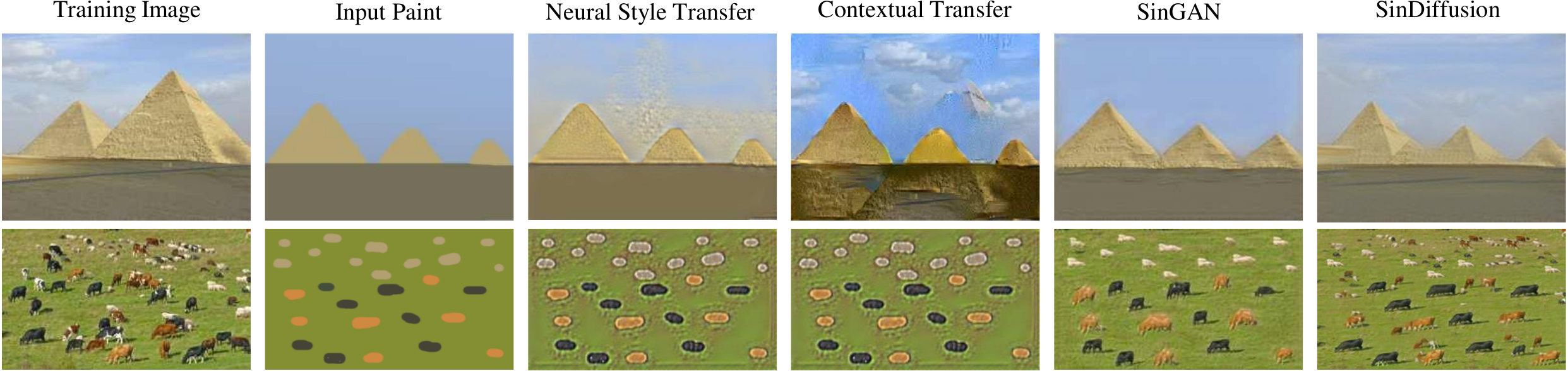}
    \vspace{-6mm}
    \caption{
    \textbf{Paint-to-image Translation.} 
    We train a SinDiffusion on a natural single image and design the sampling procedure inspired by [5].
    Our SinDiffusion generates more realistic images compared with previous methods according to the input paint image.
    }
    \vspace{-3mm}
    \label{fig:paint2img}
\end{figure*}

\vspace{2mm}
\noindent \textbf{Paint-to-Image Translation.}
The paint-to-image translation is an image-to-image translation task that aims to convert a roughly-drawing image into a photorealistic image.
Figure~\ref{fig:paint2img} shows the paint-to-image generated results of different methods.
It is observed that SinDiffusion generates more realistic and reasonable images from the paint images compared with previous methods.

\begin{figure*}[t]
    \centering
    \includegraphics[width=\linewidth]{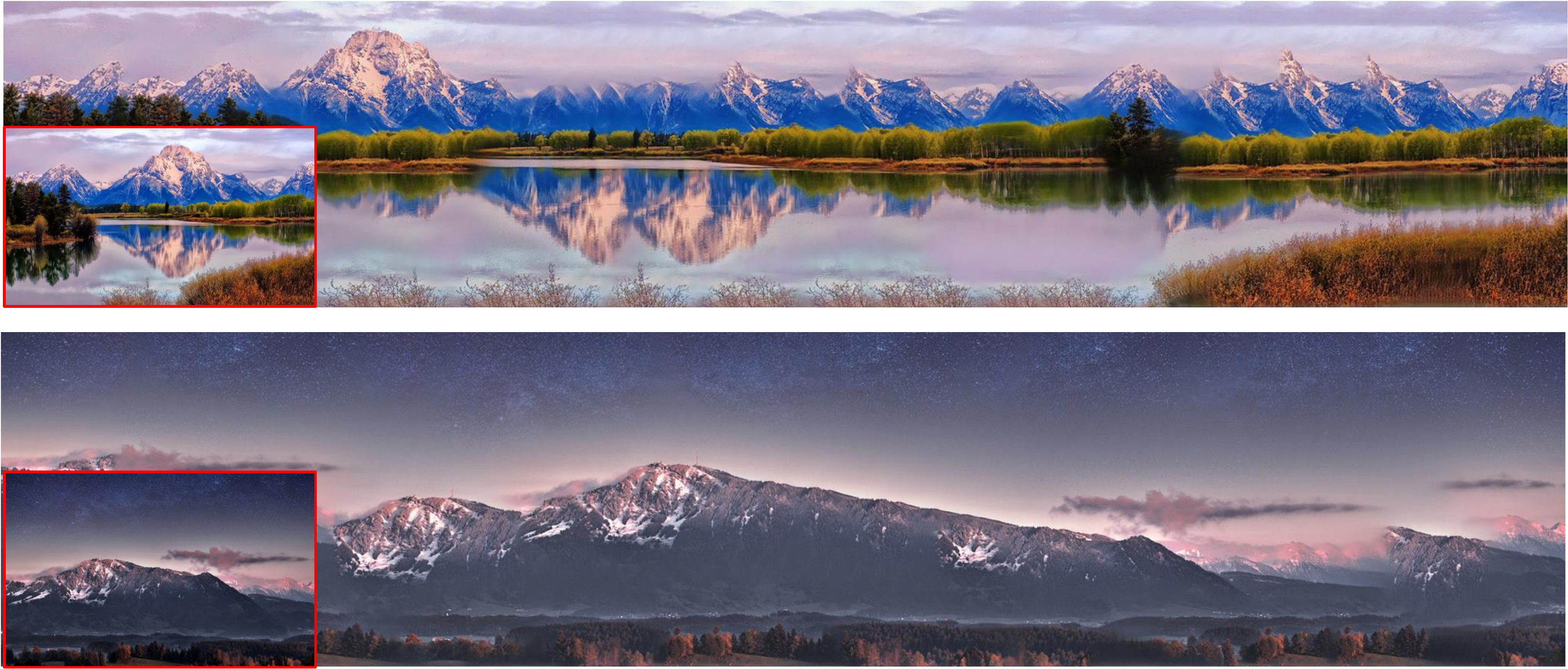}
    \vspace{-6mm}
    \caption{
    More high-resolution single image generation results with the training image.
    We utilize an enhanced SinDiffusion to accommodate high-resolution image generation.
    It is observed that our framework generates a realistic high-resolution image that contains similar patterns and structure to the training image.
    }
    \label{fig:highres}
\end{figure*}

\clearpage

\begin{figure*}[t]
    \centering
    \includegraphics[width=\linewidth]{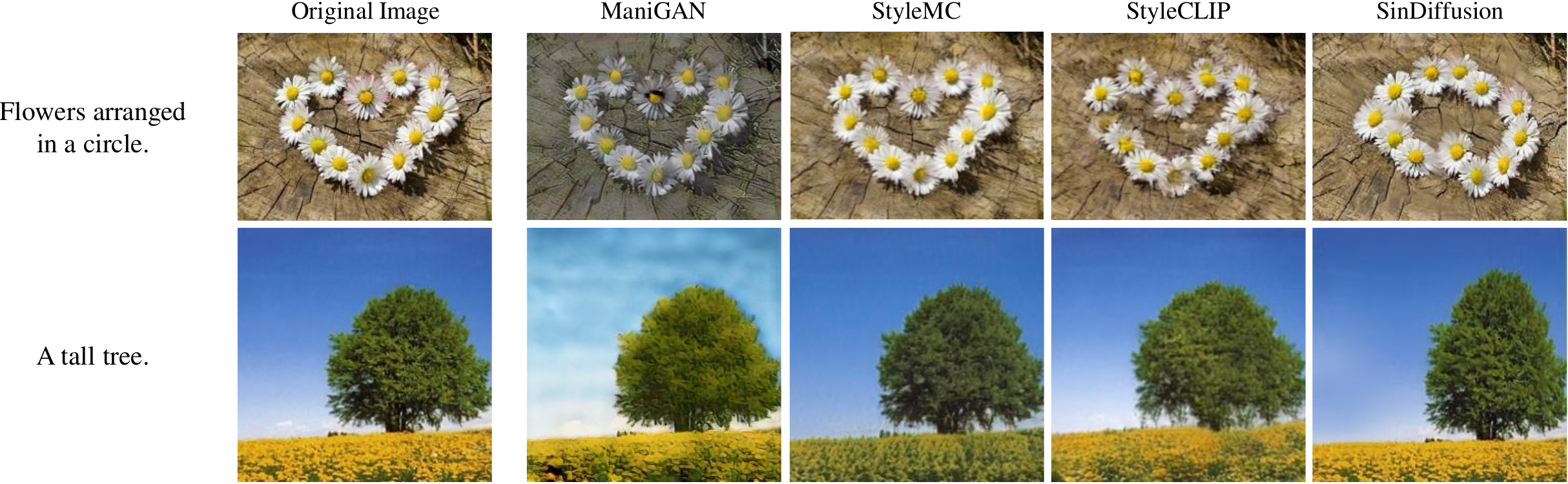}
    \vspace{-6mm}
    \caption{
    More comparison between SinDiffusion and previous methods on text-guided image generation.
    }
    \vspace{-3mm}
    \label{fig:text-guided}
\end{figure*}

\end{document}